\documentclass{bmvc2k}

\renewcommand{\thefootnote}{\fnsymbol{footnote}}

\title{HFGS: 4D Gaussian Splatting with Emphasis on Spatial and Temporal High-Frequency Components for Endoscopic Scene Reconstruction}

\addauthor{Haoyu Zhao\footnotemark[1]}{haoyu.zhao@whu.edu.cn}{1}
\addauthor{Xingyue Zhao\footnotemark[1]}{zhaoxingyue@stu.xjtu.edu.cn}{2}
\addauthor{Lingting Zhu}{ltzhu99@connect.hku.hk}{3}
\addauthor{Weixi Zheng}{acnicotine@whu.edu.cn}{1}
\addauthor{Yongchao Xu}{yongchao.xu@whu.edu.cn}{1}

\addinstitution{
 School of Computer Science,\\
 Wuhan University, \\
 Hubei, P. R. China
}
\addinstitution{
 School of Software Engineering,\\
 Xi'an Jiaotong University,\\
 Shaanxi, P. R. China
}
\addinstitution{
 School of Computing and Data Science, \\
 The University of Hong Kong,\\
 Hong Kong, P.R. China
}

\runninghead{ZHAO ET AL}{HFGS}

\usepackage{amssymb}
\usepackage{multirow}
\usepackage{xcolor}
\usepackage{colortbl}
\usepackage{bm}
\usepackage{amsmath}

\newcommand{\our}{{HFGS}\xspace}

\begin{document}
\vspace{-10pt}
{
\renewcommand{\thefootnote}{\fnsymbol{footnote}}
\footnotetext[1]{Equal contributions.}
}
\maketitle

\begin{abstract}
Robot-assisted minimally invasive surgery benefits from enhancing dynamic scene reconstruction, as it improves surgical outcomes. While Neural Radiance Fields (NeRF) have been effective in scene reconstruction, their slow inference speeds and lengthy training durations limit their applicability. To overcome these limitations, 3D Gaussian Splatting (3D-GS) based methods have emerged as a recent trend, offering rapid inference capabilities and superior 3D quality. However, these methods still struggle with under-reconstruction in both static and dynamic scenes. In this paper, we propose \textbf{\our}, a novel approach for deformable endoscopic reconstruction that addresses these challenges from spatial and temporal frequency perspectives. Our approach incorporates deformation fields to better handle dynamic scenes and introduces Spatial High-Frequency Emphasis Reconstruction (SHF) to minimize discrepancies in spatial frequency spectra between the rendered image and its ground truth. Additionally, we introduce Temporal High-Frequency Emphasis Reconstruction (THF) to enhance dynamic awareness in neural rendering by leveraging flow priors, focusing optimization on motion-intensive parts. Extensive experiments on two widely used benchmarks demonstrate that \textbf{\our} achieves superior rendering quality. Source code is available at \url{https://github.com/zhaohaoyu376/HFGS}.

\end{abstract}

\section{Introduction}
\label{sec:intro}
Endoscopic procedures are foundational to minimally invasive surgery, significantly reducing trauma and hastening patient recovery~\cite{gao2024transendoscopic,psychogyios2023sar,wang2023dynamic}. In Robotic-Assisted Minimally Invasive Surgery (RAMIS), the reconstruction of a 3D model of the surgical scene from stereo endoscopes is critical for surgical precision and efficiency. This technology enables surgeons to visualize observed tissues in 3D, which enhances their spatial awareness and navigation capabilities~\cite{mahmoud2017orbslam,wang2023video}. Despite the many benefits of endoscopic reconstruction, several challenges remain, including  limited field-of-view, obstructions, and dynamic tissue deformation~\cite{scott2008changing,zha2023endosurf,yang2023neural,wang2022neural}. Previous studies~\cite{zha2023endosurf,yang2023neural,wang2022neural,huang2024endo,liu2024endogaussian,zhu2024deformable} have successfully employed depth maps for endoscopic reconstruction; however, these methods still face two significant issues: the lack of sufficient details in generated models and inadequate rendering of non-rigid deformations.

Recent advancements in endoscopic 3D reconstruction have been significantly enhanced by Neural Radiance Fields (NeRFs)~\cite{mildenhall2021nerf}. 
EndoNeRF~\cite{wang2022neural}, a pioneering work, is the first to apply NeRF to endoscopic scenes for reconstructing deformable tissues using dual neural fields. Another approach, EndoSurf~\cite{zha2023endosurf}, utilizes the signed distance field (SDF)~\cite{yariv2020multiview,wang2021neus} to regulate surface geometry. Although these methods produce satisfactory outcomes, they demand extensive computational resources. Rendering each image necessitates querying radiance fields at numerous points and rays, which significantly limits rendering speed and poses considerable challenges for practical applications, such as intraoperative use~\cite{chen2024survey}.

\begin{figure}[h]
  \centering
  \begin{minipage}{0.325\linewidth}
    \centering
    \includegraphics[width=\linewidth]{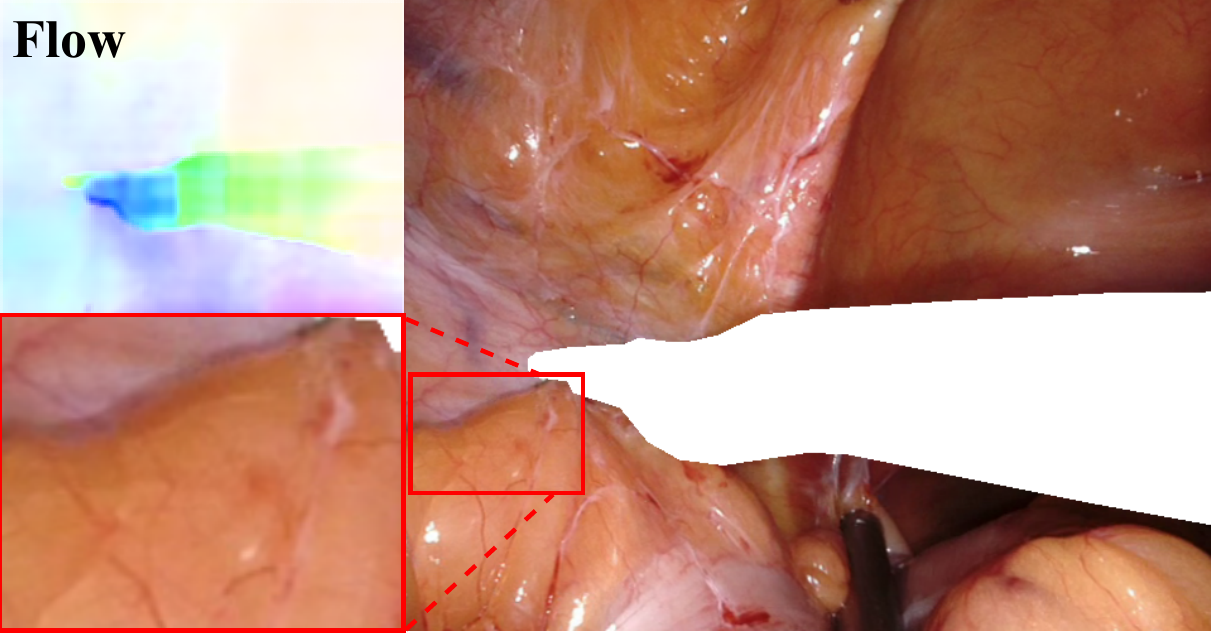}\\
    (a) GT
  \end{minipage}
  \begin{minipage}{0.325\linewidth}
    \centering
    \includegraphics[width=\linewidth]{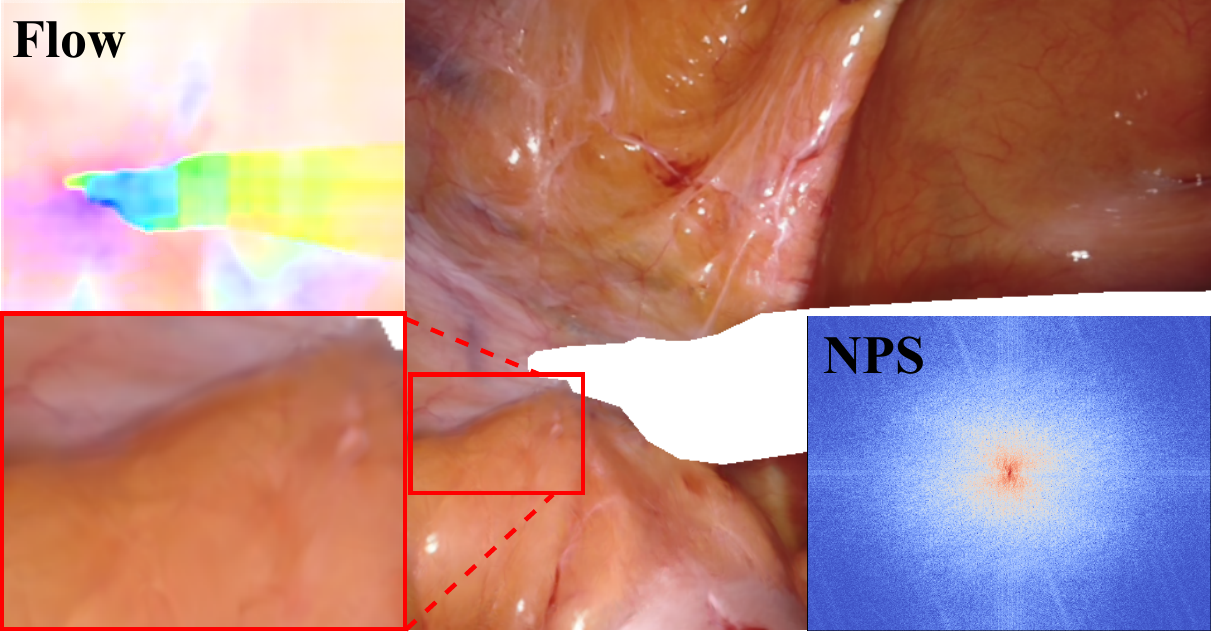}\\
    (b) EndoGS~\cite{zhu2024deformable}
  \end{minipage}
  \begin{minipage}{0.325\linewidth}
    \centering
    \includegraphics[width=\linewidth]{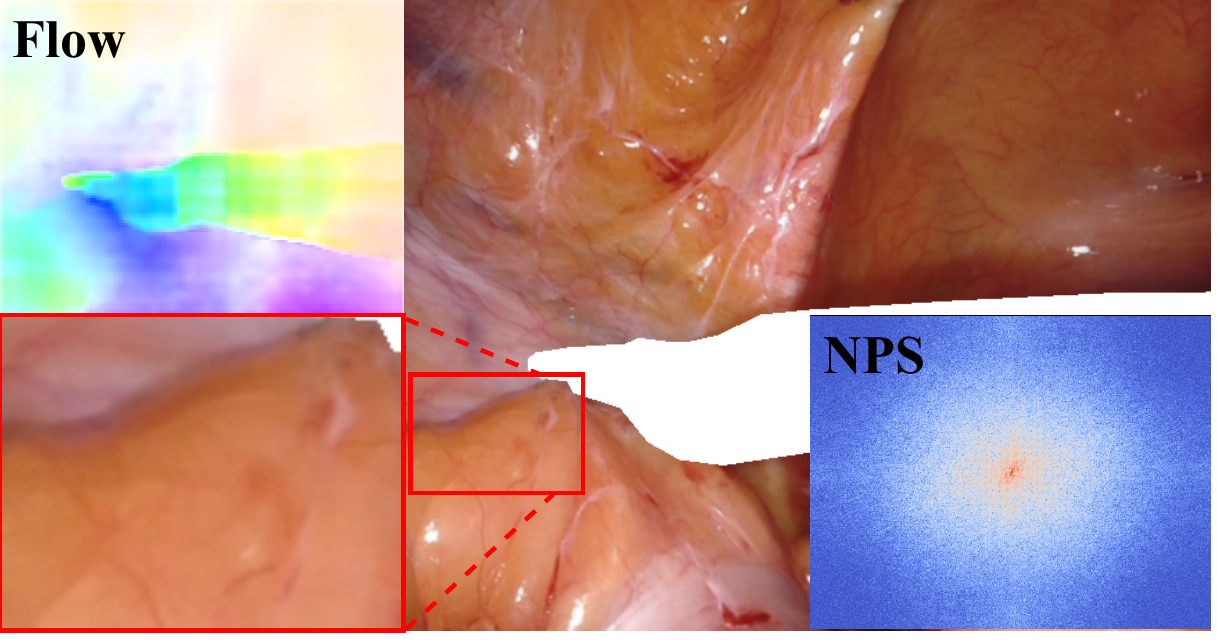}\\
    (c) \our
  \end{minipage}
  \caption{For the sample image from ENDONERF~\cite{wang2022neural}, (a-c) show the rendered image, the noise power spectrum (NPS) where blue indicates it is closer to GT, and optical flow predictions based on adjacent frame. Our \textbf{\our} not only achieves the best results, reconstructing the most detailed information and exhibiting the bluest NPS, but also renders images with optical flow that are closer to the GT.}
  \label{fig:render_compare}
\end{figure}

To address these issues, 3D Gaussian Splatting (3D-GS)~\cite{kerbl20233d} emerges as an effective alternative, providing rapid inference capability and enhanced quality of 3D representation. By optimizing anisotropic 3D Gaussians with a collection of scene images, 3D-GS effectively captures the spatial positioning, orientations, color properties, and alpha blending factors. 3D-GS reconstructs not only the geometry but also the visual texture of scenes with rapid rendering performances~\cite{kerbl20233d}. Although 3D-GS is extended to represent dynamic scenes~\cite{zhu2024deformable,liu2024endogaussian,huang2024endo,yang2023gs4d,luiten2023dynamic}, it often suffers from under-reconstruction~\cite{kerbl20233d} during the process of Gaussian densification~\cite{zhang2024fregs}, which affects both static and dynamic scenes. The under-reconstruction can be clearly observed with blur and artifacts in the rendered 2D images, the discrepancy of frequency spectrum of the render images and the corresponding ground truth, and the predicted optical flow results by~\cite{xu2023unifying}, as illustrated in Fig.~\ref{fig:render_compare}.

In this paper, we present an innovative method called \textbf{\our} for deformable endoscopic tissues reconstruction that addresses the under-reconstruction from both spatial and temporal frequency perspectives. Specifically, we propose a module called Spatial High-Frequency Emphasis Reconstruction (SHF), which minimizes the discrepancy in frequency spectra between the rendered image and the corresponding ground truth by focusing specifically on spatial high-frequency components of images. Additionally, we propose Temporal High-Frequency Emphasis Reconstruction (THF) module which enhances dynamic awareness in neural rendering by utilizing flow priors. This module targets motion areas identified through flow-based methods as temporal high-frequency components during optimization, thus improving the fidelity of moving tissues.

To summarize, our main contributions are three-fold: 1) We propose a Frequency Regularization Module to reduce spectral mismatches between rendered images and ground truth images, thereby improving in frequency space. 2) We introduce a novel module which offers dynamic awareness to existing regularization in neural rendering with the help of flow prior, providing special attention to the motion parts during optimization. 3) Experiments over multiple benchmarks show that \textbf{\our} achieves superior performances.



\section{Related Works}

\noindent
\textbf{Neural Rendering for Dynamic Scenes.}
Neural Radiance Fields (NeRF)~\cite{mildenhall2021nerf} marks a significant advance in high-quality neural rendering. Several efforts aim to adapt NeRF for dynamic scenes. For instance, some works~\cite{li2022neural,li2021neural} integrate NeRF with time-conditioned latent codes to effectively represent dynamic scenes. Another group of works~\cite{park2021nerfies,pumarola2021d,park2021hypernerf} incorporate an explicit deformation field that bends rays as they pass through various targets into a canonical space.

Some works~\cite{yang2023neural,wang2022neural,zha2023endosurf} use NeRF to represent dynamic endoscopic scenes. A good representative is EndoNeRF~\cite{wang2022neural} which follows the modeling of D-NeRF~\cite{pumarola2021d}. It trains two neural fields: one for tissue deformation and the other for canonical density and color. EndoNeRF can synthesize reasonable RGB images with post-processing filters. To tackle the lengthy training time requirement, LerPlane~\cite{yang2023neural} constructs a 4D volume by introducing 1D time to the existing 3D spatial space. Although significantly accelerating the training process, they still cannot meet the practical need of rendering speed.

\paragraph{Reconstruction with 3D Gaussian Splatting.}
Additionally, 3D-GS~\cite{kerbl20233d} is notable for its pure explicit representation and differential point-based splatting method, enabling real-time rendering of novel views through a customized CUDA-based differentiable Gaussian rasterization pipeline. The application of 3D-GS in dynamic reconstruction is just beginning to unfold. D-3DGS~\cite{luiten2023dynamic} is proposed as the first attempt to adapt 3D-GS into a dynamic setup. Other works~\cite{yang2023real, wu20234d,yang2023deformable} model 3D Gaussian motions with a compact network or 4D primitives, resulting in highly efficient training and real-time rendering.

Some works~\cite{zhu2024deformable,liu2024endogaussian} make the first attempts to apply 3D-GS to represent dynamic endoscopic scenes. EndoGS~\cite{zhu2024deformable} employs surface-aligned Gaussian Splatting~\cite{guedon2023sugar} to reconstructing deformable endoscopic tissues. EndoGaussian~\cite{liu2024endogaussian} introduces Holistic Gaussian Initialization (HGI) and Spatio-temporal Gaussian Tracking (SGT) to initialize dense Gaussians and model surface dynamics, respectively. However, these approaches often suffers from under-reconstruction~\cite{kerbl20233d} during the process of Gaussian densification~\cite{zhang2024fregs}. In contrast, our method also models 3D Gaussian motions with a deformation network for deformable endoscopic tissues reconstruction but addresses the under-reconstruction from both spatial and temporal frequency perspectives.


\section{Preliminary}
\subsection{3D Gaussian Splatting}
3D Gaussian Splatting (3D-GS)~\cite{kerbl20233d} explicitly represents scenes using point clouds, where each point is modeled as a 3D Gaussian defined by a covariance matrix $\Sigma$ and a center point $\mathcal{X}$, the latter referred to as the mean. The value at point $\mathcal{X}$ is $G(X)=e^{-\frac{1}{2}\mathcal{X}^T\Sigma^{-1}\mathcal{X}}$. For differentiable optimization, the covariance matrix $\Sigma$ is decomposed into a scaling matrix $\mathbf{S}$ and a rotation matrix $\mathbf{R}$, such that $\Sigma = \mathbf{R}\mathbf{S}\mathbf{S}^T\mathbf{R}^T$.

In rendering novel views, differential splatting as introduced by~\cite{yifan2019differentiable} and~\cite{zwicker2001surface}, involves using a viewing transform $W$ and the Jacobian matrix $J$ of the affine approximation of the projective transformation to compute the transformed covariance matrix: $\Sigma^{\prime} = JW\Sigma W^TJ^T$. Each 3D Gaussian is characterized by several attributes: position $\mathcal{X} \in \mathbb{R}^3$, color defined by spherical harmonic (SH) coefficients $\mathcal{C} \in \mathbb{R}^k$ (where $k$ is the number of SH functions), opacity $\alpha \in \mathbb{R}$, rotation factor $r \in \mathbb{R}^4$, and scaling factor $s \in \mathbb{R}^3$. The color and opacity at each pixel are computed from the Gaussian’s representation $G(X)=e^{-\frac{1}{2}\mathcal{X}^T\Sigma^{-1}\mathcal{X}}$. The blending of $N$ ordered points overlapping a pixel is given by the formula:

\begin{equation}
\label{formula: splatting&volume rendering}
    C = \sum_{i\in N}c_i \alpha_i \prod_{j=1}^{i-1} (1-\alpha_i).
\end{equation}
Here, $c_i$, $\alpha_i$ represent the density and color of this point computed by a 3D Gaussian $G$ with covariance $\Sigma$ multiplied by an optimizable per-point opacity and SH color coefficients.


\subsection{Dynamic Gaussian Splatting with Deformation Fields}
In our representation of a surgical scene as a 4-dimensional volume, the deformation of tissues is modeled over time. We adopt Gaussian deformation to represent the time-varying motions and shapes, based on the designs of~\cite{wu20234d}. Our primary objective is to accurately learn both the static parameters, $\{(\bm \mu, \bm s, \bm r, \bm{sh}, \sigma)\}$ and dynamic parameters, $\{\Delta (\bm \mu, \bm s, \bm r, \bm {sh}, \sigma)\}$ of the 3D Gaussians. For each 3D Gaussian, we compute the deformation using the the mean ${\bm \mu} = (x,y,z)$ and the time $t$. We encode the spatial and temporal information using six orthogonal feature planes~\cite{fridovich2023k, cao2023hexplane, yang2023neural, yang2023efficient, wu20234d}. Specifically, the multi-resolution HexPlane~\cite{cao2023hexplane, fridovich2023k} consists of three spatial planes $XY, XZ, YZ$ and three spatial-temporal planes $XT, YT, ZT$. These planes encode features $F \in \mathbb{R}^{h\times N_1 \times N_2}$, where $h$ represents the hidden dimension and $N_1, N_2$ indicate the plane resolution. We utilize bilinear interpolation $\mathcal{B}$ to interpolate the four nearby queried voxel features. Voxel feature can be represented in the format of matrix element-wise multiplication with operation $\odot$:

\begin{align}
f_{voxel}({\bm \mu}, t) = \mathcal{B}\left(F_{X Y}, x, y\right) \odot \mathcal{B}\left(F_{Y Z}, y, z\right) \ldots \mathcal{B}\left(F_{Y T}, y, \tau\right) \odot \mathcal{B}\left(F_{Z T}, z, \tau\right).
\end{align}

We employ a single MLP to update the attributes of Gaussian. This MLP integrates all the information to decode various parameters such as the position, scaling factor, rotation factor, spherical harmonic coefficients, and opacity:

\begin{align}
\Delta {(\bm \mu, \bm s, \bm r, \bm{sh}, \sigma)} = {\rm MLP}(f_{voxel}({\bm \mu}, t)).
\end{align}
Using the mean and time as inputs, we compute features for 3D Gaussians by querying multi-resolution voxel planes. A single MLP is then employed to derive the deformations of these Gaussians. Through differentiable rasterization, we generate rendered images and depth maps. The accuracy of these outputs is validated using ground truth images, depth maps, and tool masks, which serve as the basis for supervision.


\begin{figure}[t]
  \centering
  \includegraphics[width=1\linewidth]{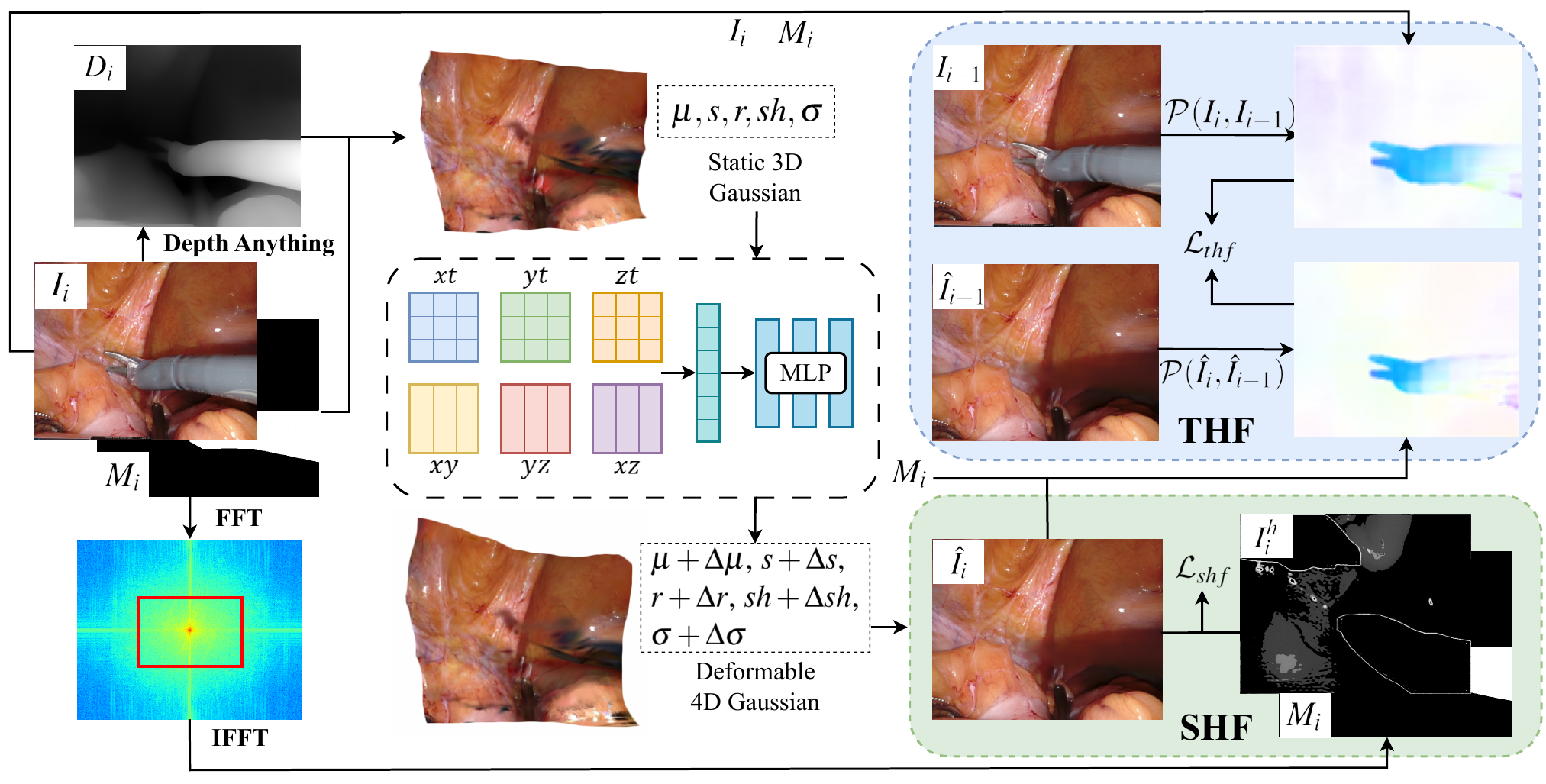}
\caption{Pipeline of the proposed \textbf{\our}. We utilize monocular images, estimated depths from Depth-Anything~\cite{yang2024depth} and tool masks for training~\cite{huang2024endo}. A single MLP is used to derive the deformation associated with these 3D Gaussian, given the features queried via voxel planes. Then we address the under-reconstruction by emphasizing spatial and temporal high-frequency components. }
   \label{fig:pipline}
\end{figure}


\section{Method}

Given a single-viewpoint stereo video of a dynamic surgical scene, we aim to reconstruct 3D structures and textures of surgical scenes without occlusion of surgical instruments. We denote ${\{(I_i, D_i, M_i)\}}^{T}_{i=1}$ as a sequence of input stereo video frames, where $T$ is the total number of frames, $I_i$$\in\mathbb{R}^{H\times W\times 3}$ and $D_i$$\in\mathbb{R}^{H\times W}$ is the $i$-th left RGB image and depth map with height $H$ and width $W$. Mask $M_i$ is utilized to specifically exclude pixels from surgical tools. Time of the $i$-th frame is $i/T$. We formulate our solution with a probabilistic model to learn statistics for depth from Depth-Anything~\cite{yang2024depth}.

In this paper, we present an innovative method called \textbf{\our} for deformable endoscopic tissues reconstruction that addresses the under-reconstruction by emphasizing spatial and temporal high-frequency components, as shown in Fig.~\ref{fig:pipline}. We first introduce Spatial High-Frequency Emphasis Reconstruction (SHF) in Section~\ref{sec:shf} which minimizes differences in the spatial frequency spectra of the rendered image and the corresponding ground truth by focusing specifically on spatial high-frequency components. We then introduce Temporal High-Frequency Emphasis Reconstruction (THF) in Section~\ref{sec:thf} which enhances dynamic awareness in neural rendering by utilizing a flow prior. This module targets motion areas identified through flow-based methods as temporal high-frequency components during optimization, thus improving the fidelity of moving tissues. Finally, we describe the optimization process in Section~\ref{sec:opt}.



\subsection{Spatial High-Frequency Emphasis Reconstruction}
\label{sec:shf}

In naive pixel-wise $L_1$ loss implementations, the average gradient might be quite small, which can occur even in regions that are not well-reconstructed, which misleads the Gaussian densification~\cite{zhang2024fregs}. As Gaussian densification is not applied to Gaussians with small gradients~\cite{kerbl20233d}, these Gaussians cannot be densified through splitting into smaller Gaussians, leading to under-reconstruction. Spatial high-frequency components focus on object structures resembling identity~\cite{1456290,zhao2024morestyle}. Thus, it is reasonable to guide the Gaussian densification by applying regularization in spatial frequency domain. For an ground truth image $I_{i}$, its frequency space signal $\mathcal{F}(I_i)$ can be obtained with Fast Fourier Transform (FFT), which is defined as follows:

\begin{equation}
\begin{aligned}
\mathcal{F}(I_i)(u,v,c) & =\sum_{h=1}^H\sum_{w=1}^WI_i(h,w,c)e^{-j2\pi(\frac hHu+\frac wWv)} =\mathcal{A}(I_i)e^{j\mathcal{P}(I_i)},
\end{aligned}
\label{eq:fft}
\end{equation}
where $j^2 = -1$, and $\mathcal{A}(I_i)$ and $\mathcal{P}(I_i)$ refer to the amplitude and phase spectra of $I_i$, respectively.
We center the low-frequency components within the frequency spectrum, and then introduce a binary mask $\mathcal{B}\in\mathbb{R}^{H\times W}$, where all values are zero except in the central region.
Following~\cite{zhao2024morestyle}, the high-frequency components $\mathcal{A}_h(I_i)$ are given by:

\begin{equation}
\begin{gathered}
\mathcal{A}_h(I_i)=(I-\mathcal{B})\odot\mathcal{A}(I_i),
I_i^h=\mathcal{F}^{-1}(\mathcal{A}_h(I_i)),
\end{gathered}
\label{eq:high_low}
\end{equation}
where $\odot$ denotes element-wise multiplication and $\mathcal{F}^{-1}$ means the Inverse Fast Fourier Transform (IFFT). We then obtain the spatial domain representation of the image $I_i^h$ that contains only the spatial high-frequency components. The $\mathcal{L}_{shf}$ is defined as:

\begin{equation}
\begin{gathered}
\mathcal{L}_{shf}(I_i,\hat{I_i})= \sum_{x=1}^{W} \sum_{y=1}^{H} I_i^h(x,y) \cdot |I_i(x,y) - \hat{I_i}(x,y)|,
\end{gathered}
\label{eq:high_low}
\end{equation}
where $\hat{I_i}$ means the rendered image.


\subsection{Temporal High-Frequency Emphasis Reconstruction}
\label{sec:thf}
To extend 3D-GS~\cite{kerbl20233d} to dynamic scenes, Gaussian motions and shape changes are modeled using a Gaussian deformation field network, as discussed in \cite{liu2024endogaussian,zhu2024deformable,wu20234d}. However, these methods struggle to effectively render dynamic images in scenarios with rapid movement, such as in 3D dynamic endoscopic scene reconstruction~\cite{guo2024motion}. This limitation stems from their insufficient use of the abundant motion data available from 2D observations. To address this, we propose a module called Temporal High-Frequency Emphasis Reconstruction (THF), which applies regularization in the temporal frequency domain of the deformation field network. This module enhances dynamic awareness in neural rendering by incorporating a flow prior. This flow prior is designed to prioritize regions exhibiting more significant movements in the current frame, thereby improving the rendering of dynamic scenes.

We feed both the rendered image $\hat{I}_i$ along with its adjacent frame $\hat{I}_{i-1}$, and the corresponding ground truth image  $I_i$ with its adjacent frame $I_{i-1}$ into a pre-trained predictor $\mathcal{P}$~\cite{xu2023unifying}. For the first frame, we treat its adjacent frame as the frame itself. This process is used to predict the optical flows $\hat{f}_i$ and $f_i$.

\begin{equation}
\begin{gathered}
\hat{f_i}=\mathcal{P}(\hat{I}_{i},\hat{I}_{i-1}),f_i=\mathcal{P}(I_{i},I_{i-1}).
\end{gathered}
\label{eq:high_low}
\end{equation}
We define the loss $\mathcal{L}_{thf}$ as the sum of the Charbonnier loss~\cite{charbonnier1994two} and the census loss~\cite{meister2018unflow}, $\mathcal{L}_{thf}=\mathcal{L}_{char}+\mathcal{L}_{cen}$, which improves the quality of interpolation, making it more resilient to outliers and structural variations in the scene.








\subsection{Optimization}
\label{sec:opt}

In reconstructing from videos with tool occlusion, we face challenges similar to~\cite{wang2022neural,yang2023neural, yang2023efficient}. We address these challenges by using labeled tool masks to identify unseen pixels. We only optimize in the seen part by introducing the term $(1-M_i)$, using the $L_1$ loss as follows:

\begin{equation}
\mathcal{L}_{L1}(I_i,\hat{I}_i) =\sum_{x=1}^{W} \sum_{y=1}^{H}|(1-M_i(x,y))\odot \hat{I}_i(x,y)-(1-M_i(x,y))\odot I_i(x,y)|.
\end{equation}


Monocular reconstruction results in limited 3D information. We address this by integrating a depth-guided loss using estimated depth maps $D_i$ with Huber loss $\mathcal{L}_D(i)$ following~\cite{zhu2024deformable}. We also apply total variation (TV) loss $\mathcal{L}_{TV}$ as~\cite{wu20234d} to regularize the rendered images. To sum up, our final optimization target is:

\begin{equation}
\begin{aligned}
    \mathcal{L}(I_i,\hat{I}_i) = 
    & \mathcal{L}_{L1}(I_i,\hat{I}_i) + \lambda_d \mathcal{L}_D(I_i,\hat{I}_i) + \lambda_{s} \mathcal{L}_{S}(I_i) + \lambda_{tv} \mathcal{L}_{TV}(I_i,\hat{I}_i) 
    \\
    & + \lambda_{shf} \mathcal{L}_{SHF}(I_i,\hat{I}_i)+\lambda_{thf} \mathcal{L}_{THF}(I_i,\hat{I}_i),
    \label{eq:R_norm}
\end{aligned}
\end{equation} 
where $\mathcal{L}_{S}$ is the surface-aligned item in EndoGS~\cite{zhu2024deformable}, which is modified from SuGaR~\cite{guedon2023sugar} to encourage the surface alignment of the Gaussians. Hyperparameters $\lambda_D$, $\lambda_{tv}$, $\lambda_{shf}$ and $\lambda_{thf}$ control the regularization strength. We set $\lambda_d$ to 0.5, $\lambda_s$ to 0.2, $\lambda_{tv}$ to 0.1, $\lambda_{shf}$ to 1, and $\lambda_{thf}$ to 10. 


\section{Experiments}

\subsection{Datasets}
We conduct experiments on two public endoscope datasets, namely ENDONERF~\cite{wang2022neural} and SCARED~\cite{allan2021stereo}. The ENDONERF dataset~\cite{wang2022neural} includes \textbf{two} cases of in-vivo prostatectomy data providing single-viewpoint estimated depth maps and manually annotated tool masks. The SCARED dataset~\cite{allan2021stereo} offers ground truth RGBD images from four porcine cadaver abdominal anatomies, using a DaVinci endoscope and a projector. We preprocess SCARED dataset according to~\cite{liu2024endogaussian}. We evaluate our method by comparing it with recent surgical scene reconstruction methods~\cite{wang2022neural,zha2023endosurf,zhu2024deformable,liu2024endogaussian} using image quality metrics such as PSNR, SSIM, and LPIPS as outlined in EndoGS~\cite{zhu2024deformable}.


\begin{table*}[h]
\centering
\setlength{\tabcolsep}{2pt} 
\begin{tabular}{l|ccc|ccc|c}
\hline
\multirow{2}{*}{Method} & \multicolumn{3}{c|}{ENDONERF} & \multicolumn{3}{c|}{SCARED} & \multirow{2}{*}{FPS}\\
 & PSNR↑ & SSIM↑ & LPIPS↓  & PSNR↑ & SSIM↑ & LPIPS↓ & \\
 \hline
EndoNeRF~\cite{wang2022neural}     & 34.20 & 0.935 & 0.156 & 23.52 & 0.754 & 0.400  & 0.2 \\
ForPlane-9k~\cite{yang2023efficient}  & 33.63 & 0.918 & 0.100 & 22.68 & 0.745 & 0.431 & 1.7 \\
ForPlane-32k~\cite{yang2023efficient}  & 36.65 & 0.947 & 0.056 & 23.50 & 0.762 & 0.348 & 1.7 \\
EndoSurf~\cite{zha2023endosurf}    & 34.99 & 0.955 & 0.113 & 23.94 & 0.779 & 0.384 & 0.04 \\
EndoGS~\cite{zhu2024deformable}            & 36.84 & 0.963 & \cellcolor{yellow}0.041 & \cellcolor{yellow}26.46 & 0.770 & \cellcolor{yellow}0.339 & \cellcolor{yellow}$\sim$70 \\
EndoGaussian~\cite{liu2024endogaussian}  & \cellcolor{yellow}37.99 & \cellcolor{yellow}0.966 & 0.043 & 26.39 & \cellcolor{yellow}0.792 & 0.530 & \cellcolor{pink}$\sim$100 \\

\our      & \cellcolor{pink}38.14 & \cellcolor{pink}0.971 & \cellcolor{pink}0.033 & \cellcolor{pink}27.47 & \cellcolor{pink}0.796 & \cellcolor{pink}0.311 & \cellcolor{yellow}$\sim$70 \\

\hline
\end{tabular}
\caption{Quantitative metrics of appearance (PSNR/SSIM/LPIPS) on ENDONERF~\cite{wang2022neural} and SCARED~\cite{allan2021stereo}. The \colorbox{pink}{best} and the \colorbox{yellow}{second best} results are denoted by pink and yellow.}
\label{tab:endonerf}
\end{table*}

\subsection{Implementation Details}

In this work, we implement a two-stage training methodology as~\cite{wu20234d}. Initially, we focus on training the static field using 3D Gaussian models, followed by training the deformation field. The training involves 3,000 iterations for the static field and extends to 60,000 for the deformation field. Initial point clouds are estimated using COLMAP \cite{schonberger2016structure}. All models are trained on an NVIDIA RTX 3090 GPU.


\begin{figure}[h]
  \centering
  \includegraphics[width=1\linewidth]{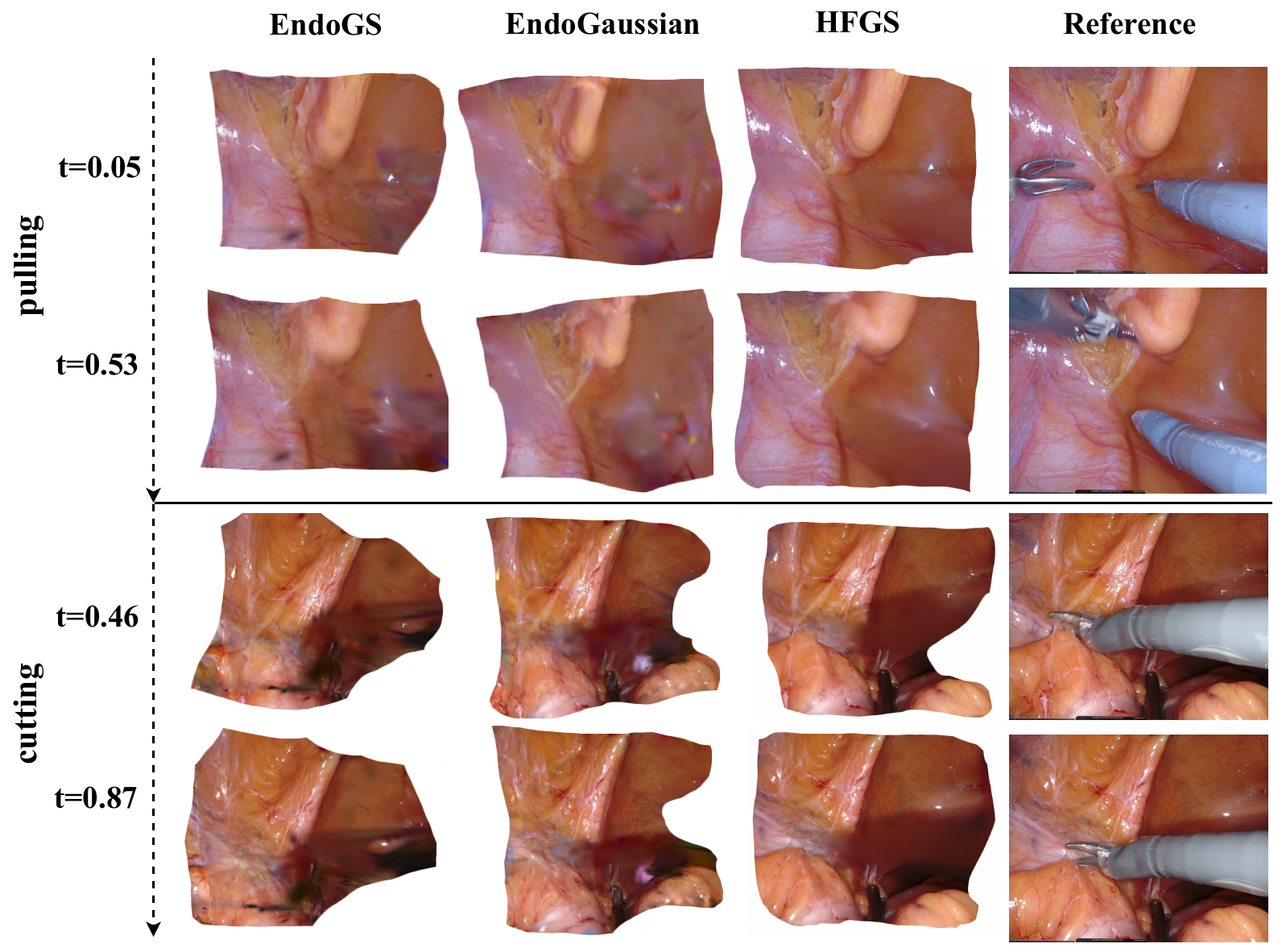}
   \caption{Illustration of reconstruction results of previous works and ours on scene "pulling soft tissues" and "cutting tissues twice" on ENDONERF~\cite{wang2022neural}.}
   \label{fig:results}
\end{figure}


\subsection{Comparison with State-of-the-art Methods}

We conduct comparative experiments against various state-of-the-art (SOTA) methods for surgical scene reconstruction, including NerF-based methods such as EndoNeRF~\cite{wang2022neural}, ForPlane~\cite{yang2023efficient} (an updated version of LerPlane~\cite{yang2023neural}) and EndoSurf~\cite{zha2023endosurf}, and 3D-GS-based methods such as EndoGS~\cite{zhu2024deformable} and EndoGaussian~\cite{liu2024endogaussian}.

Table.~\ref{tab:endonerf} presents a quantitative comparison on two public dataset. The FPS in Table.~\ref{tab:endonerf} represents the values collected by these methods on the ENDONERF~\cite{wang2022neural} dataset, where all measurements are conducted using a single NVIDIA GeForce RTX 3090 GPU. We observe that while EndoNeRF~\cite{wang2022neural}, ForPlane~\cite{yang2023efficient} and EndoSurf~\cite{zha2023endosurf} achieve high performance, however, they require hours of training and testing, making them time-consuming. In contrast, \textbf{\our} benefits from the rendering efficiency of Gaussian Splatting, enabling it to achieve real-time rendering speeds, and outperforms other SOTA methods in all evaluated metrics on both datasets.

Fig.~\ref{fig:results} presents a qualitative comparison between \textbf{\our} and competitive methods. Notably, the visualizations show that our \textbf{\our} preserves a significant amount of details with accurate geometry features. Both quantitative and qualitative results strongly support the effectiveness of \textbf{\our} in achieving high-quality 3D reconstruction at real-time inference speeds. This highlights its potential for future real-time endoscopic applications.


\begin{table*}[h]
\centering
\setlength{\tabcolsep}{3pt} 
\begin{tabular}{l|cccccc}
\hline
\multirow{2}{*}{Method} & \multicolumn{3}{c}{ENDONERF-pulling} & \multicolumn{3}{c}{ENDONERF-cutting}  \\
 & PSNR & SSIM & LPIPS  & PSNR & SSIM & LPIPS  \\
 \hline
Baseline      & 36.27 & 0.933 & 0.057 & 37.00 & 0.961 & 0.036 \\
Ours w/o SHF  & \cellcolor{yellow}38.06 & \cellcolor{yellow}0.967 & \cellcolor{yellow}0.044 & 37.51 & \cellcolor{yellow}0.969 & 0.024 \\
Ours w/o THF  & 37.93 & 0.965 & \cellcolor{yellow}0.044 & \cellcolor{yellow}37.67 & 0.968 & \cellcolor{yellow}0.023 \\
Ours        & \cellcolor{pink}38.44 & \cellcolor{pink}0.968 & \cellcolor{pink}0.043 & \cellcolor{pink}37.83 & \cellcolor{pink}0.969 & \cellcolor{pink}0.022 \\

\hline
\end{tabular}
\caption{Ablation studies on the impact of each module in our method on ENDONERF~\cite{wang2022neural}.}
\label{tab:ablation}
\end{table*}


\subsection{Ablation Studies}

\begin{figure}[h]
  \centering
  \begin{minipage}{0.325\linewidth}
    \centering
    \includegraphics[width=\linewidth]{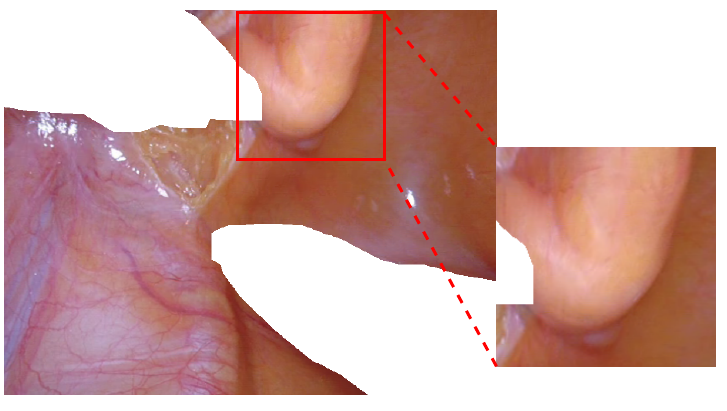}\\
    (a) Reference
  \end{minipage}
  \begin{minipage}{0.325\linewidth}
    \centering
    \includegraphics[width=\linewidth]{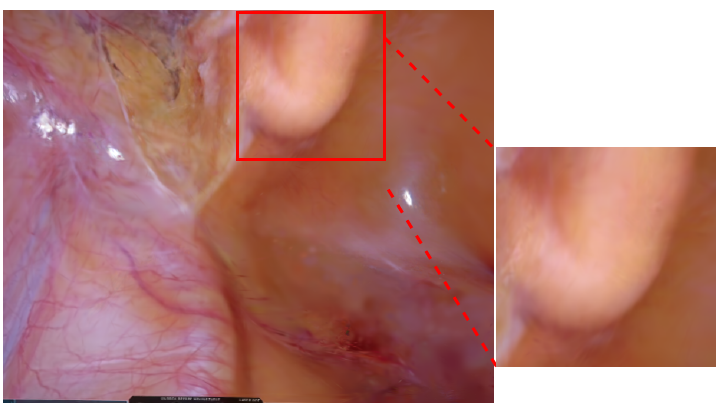}\\
    (b) w/o SHF
  \end{minipage}
  \begin{minipage}{0.325\linewidth}
    \centering
    \includegraphics[width=\linewidth]{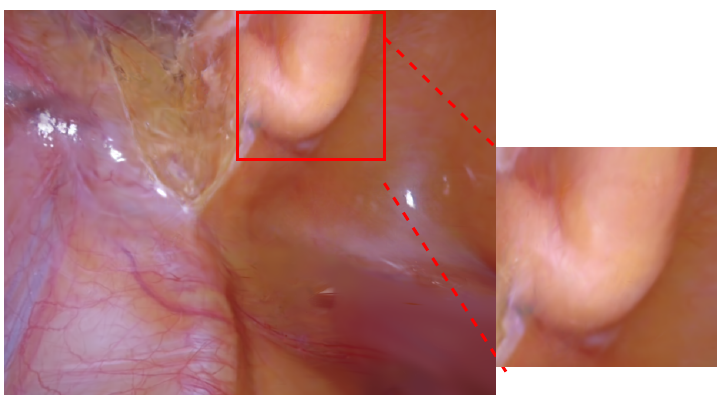}\\
    (c) w/ SHF
  \end{minipage}
  \caption{Ablation on SHF. We show rendering frames w/ and w/o SHF on scene "pulling soft tissues" on ENDONERF~\cite{wang2022neural}.}
  \label{fig:ablation}
\end{figure}

To evaluate the effectiveness of our proposed modules, including SHF and THF, we conduct ablation experiments using the ENDONERF~\cite{wang2022neural} dataset. The corresponding results are shown in Table~\ref{tab:ablation}. In Fig.~\ref{fig:ablation}, we show the effectiveness of the THF. THF helps the model reconstruct more detailed information and addresses the under-reconstruction in static scenes. Baseline method struggles to effectively render dynamic images. This phenomenon can be mitigated during the optimization with THF as shown in Fig.~\ref{fig:render_compare}. Results with THF are closer to the ground truth in the predicted optical flow results by~\cite{xu2023unifying}, indicating more accurate rendering of  dynamic scenes.

To sum up, SHF and THF all contribute to performance gains and address the under-reconstruction both in static and dynamic scenes. It is worth mentioning that our \textbf{\our} has a huge improvement based on the baseline \textbf{without any extra inference time}.


\section{Conclusion}

We introduce a method for deformable endoscopic tissue reconstruction that leverages spatial and temporal frequency analyses to improve under-reconstruction issues, enabling high-quality, real-time reconstruction from single-viewpoint videos. Our method includes two modules that enhance rendering in both static and dynamic scenes. Testing on two public datasets confirms significant performance gains over existing methods. However, 3D reconstruction from single-viewpoint videos still faces challenges for surgical use. Future work should focus on integrating multiple surgical cameras to enhance 3D tissue reconstruction accuracy and practicality in clinical environments.

\bibliography{egbib}
\end{document}